\RequirePackage{fix-cm}
\documentclass[a4paper,11pt]{article}

\usepackage{authblk}
\usepackage{algorithm}
\usepackage{algorithmic}
\usepackage{multirow}
\usepackage{graphicx}
\usepackage{booktabs}
\usepackage{epsfig}
\usepackage{epstopdf}
\usepackage{amsmath}
\usepackage{caption}
\usepackage{subcaption}
\usepackage{amsfonts}
\usepackage{wasysym}
\usepackage{paralist}
\usepackage{url}
\usepackage{float}
\usepackage{booktabs}
\usepackage{microtype}

\providecommand{\keywords}[1]{\textbf{\textit{Keywords }} #1}

\begin{document}

\title{Modeling Strong Physically Unclonable Functions with Metaheuristics}

\author[1]{Carlos Coello Coello}
\author[2]{Marko Djurasevic}
\author[2]{Domagoj Jakobovic}
\author[3]{Luca Mariot}
\author[3]{Stjepan Picek}
	
\affil[1]{{\small CINVESTAV-IPN, Departamento de Computaci\'{o}n, Mexico City, Mexico}

    {\small \texttt{ccoello@cs.cinvestav.mx}}}

\affil[2]{{\small Faculty of Electrical Engineering and Computing, University of Zagreb, Unska 3, Zagreb, Croatia}
	
	{\small \texttt{\{marko.durasevic,domagoj.jakobovic\}@fer.hr}}}

\affil[3]{{\small Digital Security Group, Radboud University, PO Box 9010, 6500 GL Nijmegen, The Netherlands} 
	
	{\small \texttt{\{luca.mariot,stjepan.picek\}@ru.nl}}}

\maketitle

\begin{abstract}
Evolutionary algorithms have been successfully applied to attacking Physically Unclonable Functions (PUFs). CMA-ES is recognized as the most powerful option for a type of attack called the reliability attack. While there is no reason to doubt the performance of CMA-ES, the lack of comparison with different metaheuristics and results for the challenge-response pair-based attack leaves open questions if there are better-suited metaheuristics for the problem.

In this paper, we take a step back and systematically evaluate several metaheuristics for the challenge-response pair-based attack on strong PUFs. Our results confirm that CMA-ES has the best performance, but we also note several other algorithms with similar performance while having smaller computational costs. More precisely, if we provide a sufficient number of challenge-response pairs to train the algorithm, various configurations show good results. Consequently, we conclude that EAs represent a strong option for challenge-response pair-based attacks on PUFs.
\end{abstract}

\keywords{Metaheuristics, Physically Unclonable Functions, CMA-ES, Challenge-response Pairs}

\section{Introduction}
\label{sec:introduction}

Evolutionary algorithms (EAs) encompass various metaheuristic techniques successfully applied to many problems, e.g., scheduling~\cite{DURASEVIC2016419}, transportation~\cite{BAKER2003787}, bioinformatics~\cite{Manning13}, and cybersecurity~\cite{10.1145/3377929.3389886}.
Again, we can find numerous examples within the cybersecurity domain where evolutionary algorithms represent a powerful paradigm. Some noteworthy examples include the design of cryptographic primitives~\cite{10.1145/3449639.3459362}, network security~\cite{10.1145/3067695.3076081}, fuzzing~\cite{rawat_vuzzer_2017}, adversarial examples~\cite{onepixel}, and attacking physically unclonable functions~\cite{10.1145/1866307.1866335}.

Physically Unclonable Functions (PUFs) are (partly) disordered physical systems that can be challenged with external stimuli upon which they react with the corresponding responses. Those responses will depend on the nanoscale structural disorder present in the PUFs.
When supplied with the same challenge, no two PUFs will give the same response. 
As such, PUFs represent a cost-efficient replacement technology for secure non-volatile memory (NVM), and they have found their applicability in modern security.

While the name suggests that a PUF cannot be cloned, numerous results have shown that various artificial intelligence techniques could easily model its behavior. More precisely, machine learning and metaheuristics found their place in the mathematical modeling of PUFs.
Machine learning is mainly used for modeling attacks based on challenge-response pairs, while metaheuristics have been found to be more successful for the so-called reliability attacks where it is required to account for the randomness in the responses.

Interestingly, while metaheuristics have been very useful in attacking PUFs, limited results are available regarding their performance for challenge-response pairs attacks. No results are available regarding the performance of metaheuristics in this problem or the difficulty in tuning them.

\subsection{Previous Related Work}

When evaluating the literature related to EAs and PUFs, we can informally classify it into two perspectives: 1) the security perspective, where the main goal is to present an algorithm that works well, i.e., manages to model a PUF, and 2) the evolutionary perspective, where the goal is to assess different algorithms and compare their performance.

The first perspective provides significantly more results. Rührmair et al. provided the first results showing it is possible to model PUFs~\cite{10.1145/1866307.1866335}. The authors evaluate ``various machine learning techniques, including Logistic Regression and Evolution Strategies.'' They showed that both techniques work well, but with evolution strategies, they managed to break specific problem instances that were not breakable with logistic regression.
In the following years, the Covariance Matrix Adaptation Evolution Strategy (CMA-ES)~\cite{Hansen2006} became the dominant option for attacking PUFs with EAs, especially concerning the reliability attack.
For instance, Becker used CMA-ES to break a specific type of PUFs called XOR PUF, where the author managed to break commercial PUF-based RFID tags~\cite{10.1007/978-3-662-48324-4_27}.
Interestingly, while CMA-ES performed well, there was little research investigating whether other metaheuristics perform comparably or how difficult it is to tune such algorithms. From a metaheuristics perspective, research on PUFs has been rare. We are aware of only a few works, but they all consider attacking PUFs as an interesting benchmark problem, rather than as the underlying research goal. For instance, Picek et al. proposed to use the problem of attacking PUFs as a benchmark problem for various EAs~\cite{10.1145/3067695.3082535}. Similarly, Picek et al. considered attacking PUFs as a benchmark problem to evaluate the influence of genotype size on evolutionary algorithms' performance~\cite{10.1145/3449726.3463169}.

\subsection{Our Contributions}

The main contributions of our work are:
\begin{compactenum}
\item We systematically evaluate six metaheuristics and seven different PUF problem instances. Our results show that the CMA-ES performs best over the tested instances.
\item When considering Arbiter PUFs, our results show that multiple algorithms work well. For XOR Arbiter PUFs, the problem is more challenging, and we observe good results with CMA-ES only (provided that sufficient challenge-response pairs are given).
\item Our results indicate that parameter tuning plays a significant role for most algorithms, and improper tuning results in poorer prediction accuracy. Interestingly, CMA-ES shows robustness, with tuning giving only small performance differences.
\item Provided there are enough challenge-response pairs in the training set, we observe a generalization effect: if the model works on the training set, it will also work on the test set.
\end{compactenum}

\section{Physically Unclonable Functions}
\label{sec:background}

Physically Unclonable Functions (PUFs) are lightweight hardware devices commonly used in authentication schemes and anti-coun\-terfeiting applications. PUFs use inherent manufacturing differences within every physical object to give each instance a unique identity. 
PUFs are usually divided into two categories: weak PUFs and strong PUFs. 
A strong PUF can be queried with an exponential number of challenges to receive an exponential number of responses (challenge-response pairs - CRP). They are used in authentication protocols and for key generation and storage. Weak PUFs have a limited challenge space (polynomial in the PUF area) and can only be used for key generation and storage.
Strong PUFs have a large challenge space, enabling them to be used in challenge-response protocols where expensive traditional cryptographic primitives cannot.
Commonly, strong PUF responses are directly available to an attacker. In the standard threat model, it is either assumed that an attacker can passively eavesdrop on many protocol runs or query responses directly from the PUF for arbitrary challenges.

Existing strong PUFs can be simulated in software, and the required parameters for such a software model can be approximated by using machine learning or evolutionary algorithms~\cite{10.1007/978-3-662-48324-4_27, 10.1145/3067695.3082535}. The attacker's goal is to build a clone of a strong PUF that generates correct responses for arbitrary challenges. 
Usually, strong PUFs rely on delayed-based Arbiter PUFs (APUFs) as their primary building blocks for PUF constructs and protocols~\cite{7450665}. Such APUFs can be modeled by a linear function, which is at the foundation of various AI-based attacks using challenge-response pairs~\cite{DBLP:journals/tifs/Delvaux19}. 

\subsection{Arbiter PUFs}

\begin{figure}[t]
     \centering
         \includegraphics[width=\columnwidth]{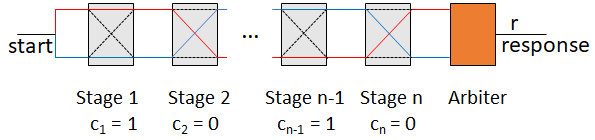}
         \caption{Depiction of an APUF.}
         \label{fig:puf}
\end{figure}

Arbiter PUFs (APUFs) consist of one or more chains of 2-bit multiplexers pairs with identical layouts.
Each multiplexer pair is denoted as a stage, with $n$ stages in a single chain.
A single input signal is introduced to the first stage to both the bottom and top multiplexer in the pair.
The chain is fed a control signal of $n$ bits called a challenge, where each bit determines whether the two input signals in that stage would be switched (crossed over by the multiplexer) or not.
In ideal conditions, the input signal would propagate at the same speed through each stage, and both the lower and upper signals would arrive at the arbiter (at the end of the chain) simultaneously.
Due to the manufacturing inconsistencies, each multiplexer delay is slightly different, and the top and bottom input signals are not synchronized.
At the end of the chain, the arbiter determines which signal arrived earlier and thus forms the response (0 or 1).
The response of a PUF is determined by the delay difference between the top and bottom input signal, which is, in turn, the sum of delay differences of the individual stages.
We depict an APUF in Figure~\ref{fig:puf}. Note that an APUF with $n$ challenges consists of $n+1$ stages as the last stage is due to the arbiter.

\subsection{XOR APUF}

\begin{figure}[t]
     \centering
         \includegraphics[width=0.8\columnwidth]{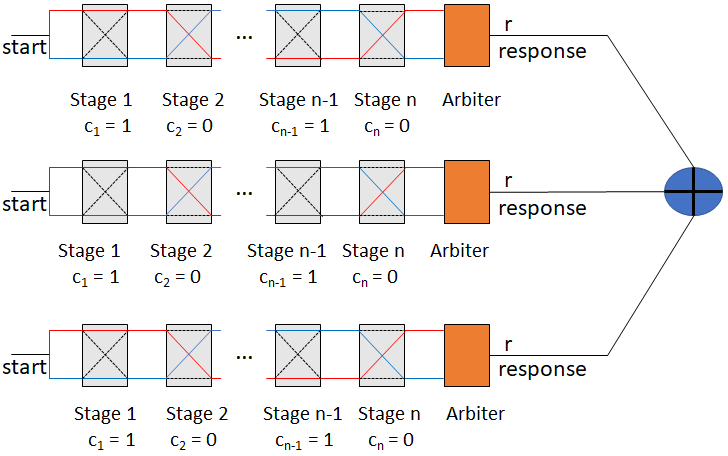}
         \caption{Depiction of a $3$-XOR APUF.}
         \label{fig:xor_puf}
\end{figure}

To increase the resistance of APUFs against machine learning attacks, it is possible to add nonlinear elements to the PUF design. One standard method is the so-called XOR APUF design~\cite{4261134}.
In a $k$-XOR APUF, $k$ Arbiter PUFs are placed on the chip. Each of the Arbiter PUFs receives the same challenges, and the responses of the $k$ PUFs are XORed to build the final response bits. With this approach, the machine learning resistance increases by XORing more PUFs, but adding additional PUF instances will increase the area overhead of the design. What is more, the XOR PUFs become more unreliable as more PUFs are XORed. Hence, there will be a limit to the number of XORs used in practice. We depict a $3$-XOR APUF in Figure~\ref{fig:xor_puf}.

\subsection{Attacking the PUFs}

There are multiple successful attacks on various strong PUFs, that can be divided into CRP-based attacks and reliability attacks. Note that it is often not necessary to reach very high accuracy, and any prediction accuracy significantly higher than 50\% can be considered as a successful attack~\cite{DBLP:journals/tifs/Delvaux19}.

\paragraph{CRP-based Attack}

To efficiently model a PUF, one tries to determine the delay vector $w=(w_1,\ldots,w_{n+1})$ that models the delay differences in each stage. 
Lim et al. proposed a linear additive model that captures the APUF behavior where we require the map $f(c) = \phi$ of the applied challenge $c$ of length $n$ to a feature vector $\phi$ of length $n + 1$~\cite{1561249}. The product of the feature vector and delay vector decides which signal came first, and based on this, what is the response bit $r$:
\begin{eqnarray}
\phi_i = \prod_{l=i}^{k}(-1)^{c_l}, \text{for } 1 \leq i \leq k.\\ 
r = \begin{cases}
1 &\text{if \  $\vec{w}\vec{\phi}^T < 0$}\\
0 &\text{if \ $\vec{w}\vec{\phi}^T > 0$} 
\end{cases}
\end{eqnarray}

The research community explored various algorithms for modeling attacks, both from the EA and machine learning domains. For more details, we refer interested readers to~\cite{DBLP:phd/basesearch/Delvaux17}.

\paragraph{Reliability Attack}

As already stated, attacking XOR APUFs is more difficult than attacking APUFs due to added nonlinearity. Further, the more APUFs are in a XOR APUF, the more the attack becomes difficult. 
To make the attack easier, it is possible also to consider the reliability of a response, i.e., how often the PUF evaluates to the same response bit for a given challenge~\cite{6581579}~\footnote{While using reliability, the attack in~\cite{6581579} did not apply it to XOR APUFs.}.
The main idea of the reliability attack is to make repeated measurements for the same challenge and observe which response bits are stable and which response bits sometimes flip. Then, if the response for a given challenge is unstable, it is likely that the corresponding delay difference is close to zero~\cite{10.1007/978-3-662-48324-4_27}.
The reliability attacks are often conducted with EAs and especially with CMA-ES.

\section{Methodology}
\label{sec:methodology}

\subsection{Experimental Design}

In a CRP-based attack, the attacker tries to infer the delay vector that adequately describes the PUF under consideration. The obvious choice for the solution encoding is a floating-point vector that models the target device behavior. For an APUF, the solution size is equal to $n+1$ where $n$ is the number of stages, while the solution for a XOR APUF will have $k \times (n+1)$ floating-point values.

From the attacker's perspective, the optimization goal is to predict as many correct responses as possible, given a set of challenges. 
The model optimization is performed on the learning set of challenge-response pairs (CRPs), while the generalization capability is evaluated on a separate test set.
Therefore, our simple fitness function minimizes the number of errors, that is, wrongly predicted responses: $fitness = $ (number of prediction errors).

The attacks are simulated on a number of PUF sizes in increasing complexity; the first group of experiments is conducted on APUFs in sizes of 16, 32, 64, and 128 stages.
The second group considers XOR APUFs of sizes $4 \times 16$, $4 \times 32$, and $4 \times 64$, which are commonly found in available devices.

The \textit{learning set} for every PUF size is devised in the following manner: we randomly create ten independent PUFs and generate corresponding learning sets with varying numbers of CRPs, ranging from 2\,000 to 250\,000.
Every algorithm is applied to each of the ten instances in five runs, which results in 50 independent runs.
The reason for using ten different PUF instances is to reduce the bias that could arise from experimenting on a single PUF only.
All the PUFs are instantiated by randomly generating their delay vectors using a normal distribution with parameters $\mathcal{N}(0, 1)$, following the literature.

The test set is produced with the same PUF instances used in the learning phase but with different challenge-response pairs.
Each PUF instance will have a single test set, and solutions obtained on all five runs will be tested on that same set.
Unlike the learning sets, all the test sets have the same size of 1\,000 CRPs.

\subsection{Algorithms and Parameter Tuning}

We applied the following algorithms to this problem: artificial immune system algorithm (AIS)~\cite{Bernardino2009}, clonal selection algorithm (CLONALG)~\cite{Castro2002}, covariance matrix adaptation ES (CMA-ES)~\cite{hansen2004}, differential evolution (DE)~\cite{Storn1997}, a generational genetic algorithm with roulette wheel selection (RW), and a steady-state genetic algorithm with tournament selection (SST)~\cite{Goldberg1989}. Besides these algorithms, evolution strategy (ES)~\cite{Beyer2002}, particle swarm optimization (PSO)~\cite{Kennedy1995}, artificial bee colony (ABC)~\cite{Karaboga2005}, and elimination genetic algorithm~\cite{jakobovic97} were also tested. However, due to their poor performance during the parameter tuning procedure, they were not considered in further experiments. 
The algorithmic descriptions are omitted to save space, but their definitions and available parameters can be found at the library website: \url{http://ecf.zemris.fer.hr/}

Prior to comparison, each algorithm was tuned on a small set of PUF instances to estimate optimal parameter values; the tuning was based on learning accuracy only. The parameter values for each selected algorithm after tuning are shown in Table~\ref{tbl:params}. All algorithms used the same termination criterion of 100\,000 function evaluations to ensure a fair comparison. 

\begin{table}[!h]
\caption{Selected parameter values.}
\label{tbl:params}
\centering
\begin{tabular}{@{}ll@{}}
\toprule
Parameter                                                       & Value \\ \midrule
Artificial immune algorithm (AIS)                               &       \\ \midrule
Population size                                                 & 50    \\
Mutation rate                                                   & 0.1   \\
Number of clones                                                & 5     \\
Number of generations to keep an individual                     & 68    \\ \midrule
Clonal selection algorithm (CLONALG)                            &       \\ \midrule
Population size                                                 & 100   \\
Mutation rate                                                   & 0.2   \\
Cloned antibodies in each generation                            & 10    \\
Number of clones for every antibody (\%)                        & 0.9   \\
Fraction of population which is regenerated                     & 0     \\ \midrule
Covariant matrix adapation (CMA-ES)                             &       \\ \midrule
Population size                                                 & 20    \\
Mu                                                              & 3     \\ 
Mutation probability                                            & 0.7   \\ \midrule
Differential evolution (DE)                                     &       \\ \midrule
Population size                                                 & 500   \\
Mutation probability                                            & 0.5   \\
Scaling constant                                                & 0.4   \\
Crossover rate                                                  & 0.9   \\ \midrule
Roulette wheel genetic algorithm (RW)                           &       \\ \midrule
Population size                                                 & 500   \\
Mutation probability                                            & 0.5   \\
Crossover probability                                           & 0.9   \\
Selection pressure                                              & 20    \\ \midrule
Steady state genetic algorithm (SST)                            &       \\ \midrule
Population size                                                 & 20    \\
Mutation probability                                            & 0.9   \\
Tournament size                                                 & 3     \\ \bottomrule
\end{tabular}
\end{table}

\section{Experimental Results}
\label{sec:results}
This section starts with results for various instances of APUFs. Afterwards, we provide experimental results for various instances of XOR PUFs. Finally, we provide several general observations.

\subsection{Results for APUFs}

Table~\ref{tbl:apuf} shows the errors obtained for different APUF instances on the test set by different optimization algorithms. The table lists only the lowest error value obtained by each method (the whole range of error values is depicted below using violin plots).
We note that all methods achieve good results for the simplest PUF instance, with the error mostly below 1\%. As the complexity of the PUF instance increases, the errors of all methods also increase. However, with the increase in the number of CRPs, it is possible to decrease this error for all algorithms to some extent. Among the tested algorithms, CMA-ES achieved the best results in all experiments, as it achieved an error of 0 for all tested problem instances. Among the other algorithms, AIS, CLONALG, and SST performed competitively. Because all algorithms are capable of reaching more than 80\% prediction accuracy even on the instance with the smallest training set, we conclude that APUFs can indeed be efficiently broken with EAs.

\begin{table}[!h]
\caption{Test set errors by different APUF instances.}
\label{tbl:apuf}
\centering
\begin{tabular}{@{}llcccccc@{}}
\toprule
\multirow{2}{*}{PUF}  & \multirow{2}{*}{CRP} & \multicolumn{6}{c}{Method}                                                                                                                                    \\ \cmidrule(l){3-8} 
                       &                     & \multicolumn{1}{l}{AIS} & \multicolumn{1}{l}{CLONALG} & \multicolumn{1}{l}{CMA-ES} & \multicolumn{1}{l}{DE} & \multicolumn{1}{l}{RW} & \multicolumn{1}{l}{SST} \\ \midrule
\multirow{3}{*}{1$\times$16}  & 2\,000                & 0                       & 3                           & 1                         & 0                      & 12                     & 1                       \\
                       & 10\,000               & 0                       & 0                           & 0                         & 1                      & 12                     & 0                       \\
                       & 50\,000               & 0                       & 0                           & 0                         & 0                      & 10                     & 0                       \\ \midrule
\multirow{3}{*}{1$\times$32}  & 2\,000                & 12                      & 16                          & 3                         & 8                      & 30                     & 12                      \\
                       & 10\,000               & 1                       & 3                           & 0                         & 8                      & 21                     & 3                       \\
                       & 50\,000               & 0                       & 0                           & 0                         & 5                      & 19                     & 0                       \\ \midrule
\multirow{3}{*}{1$\times$64}  & 2\,000                & 43                      & 85                          & 8                         & 37                     & 36                     & 64                      \\
                       & 10\,000               & 12                      & 10                          & 0                         & 30                     & 35                     & 11                      \\
                       & 50\,000               & 6                       & 1                           & 0                         & 30                     & 32                     & 3                       \\ \midrule
\multirow{3}{*}{1$\times$128} & 2\,000                & 156                     & 193                         & 35                        & 87                     & 94                     & 174                     \\
                       & 10\,000               & 50                      & 32                          & 5                         & 73                     & 67                     & 43                      \\
                       & 50\,000               & 22                      & 12                          & 0                         & 69                     & 65                     & 14                      \\ \bottomrule
\end{tabular}
\end{table}

\begin{figure}[!h]
     \centering
     \begin{subfigure}[t]{0.5\textwidth}
         \centering
         \includegraphics[width=\columnwidth]{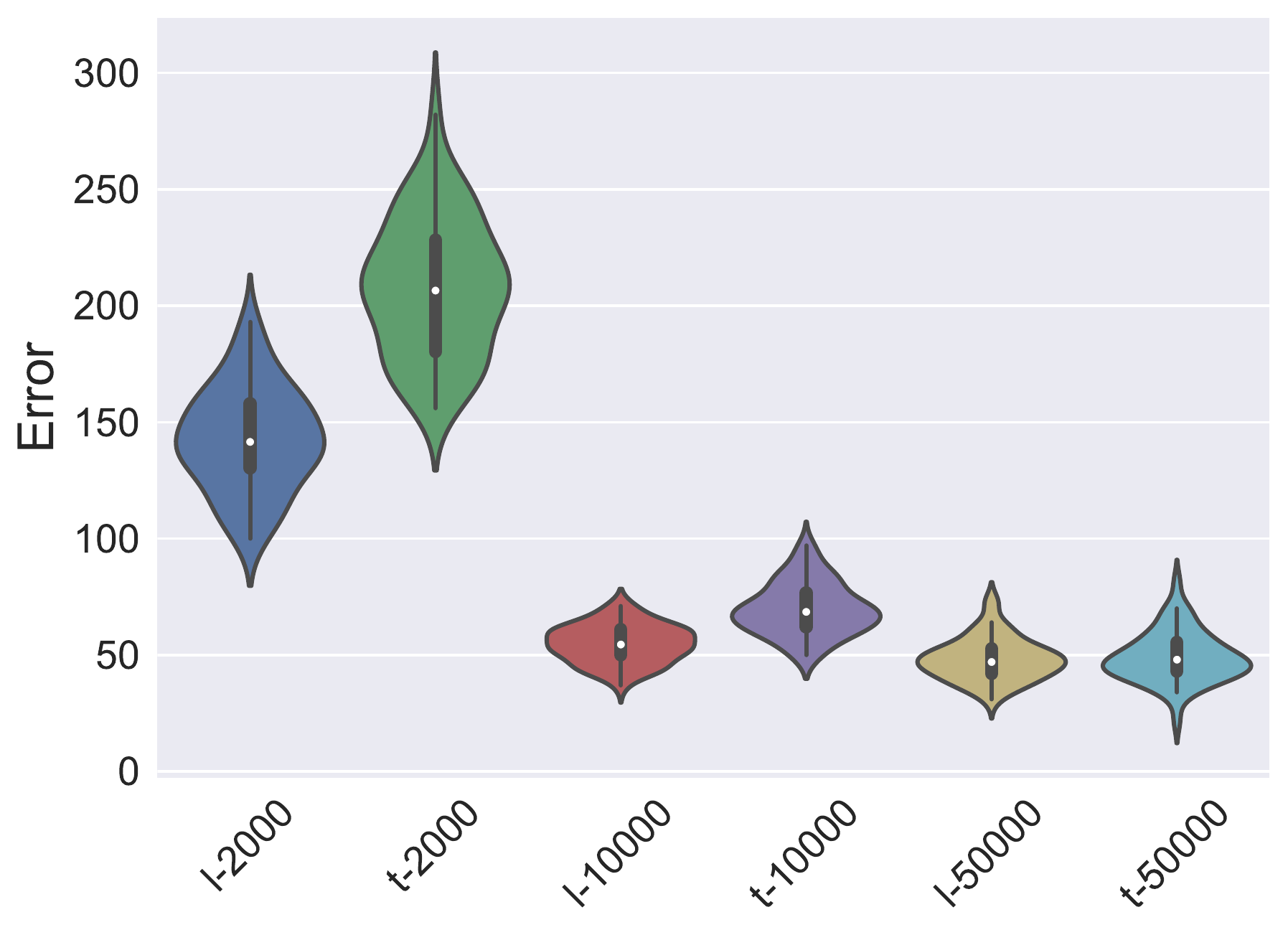}
         \caption{AIS.}
         \label{fig:y equals x1}
     \end{subfigure}%
     \begin{subfigure}[t]{0.5\textwidth}
         \centering
         \includegraphics[width=\columnwidth]{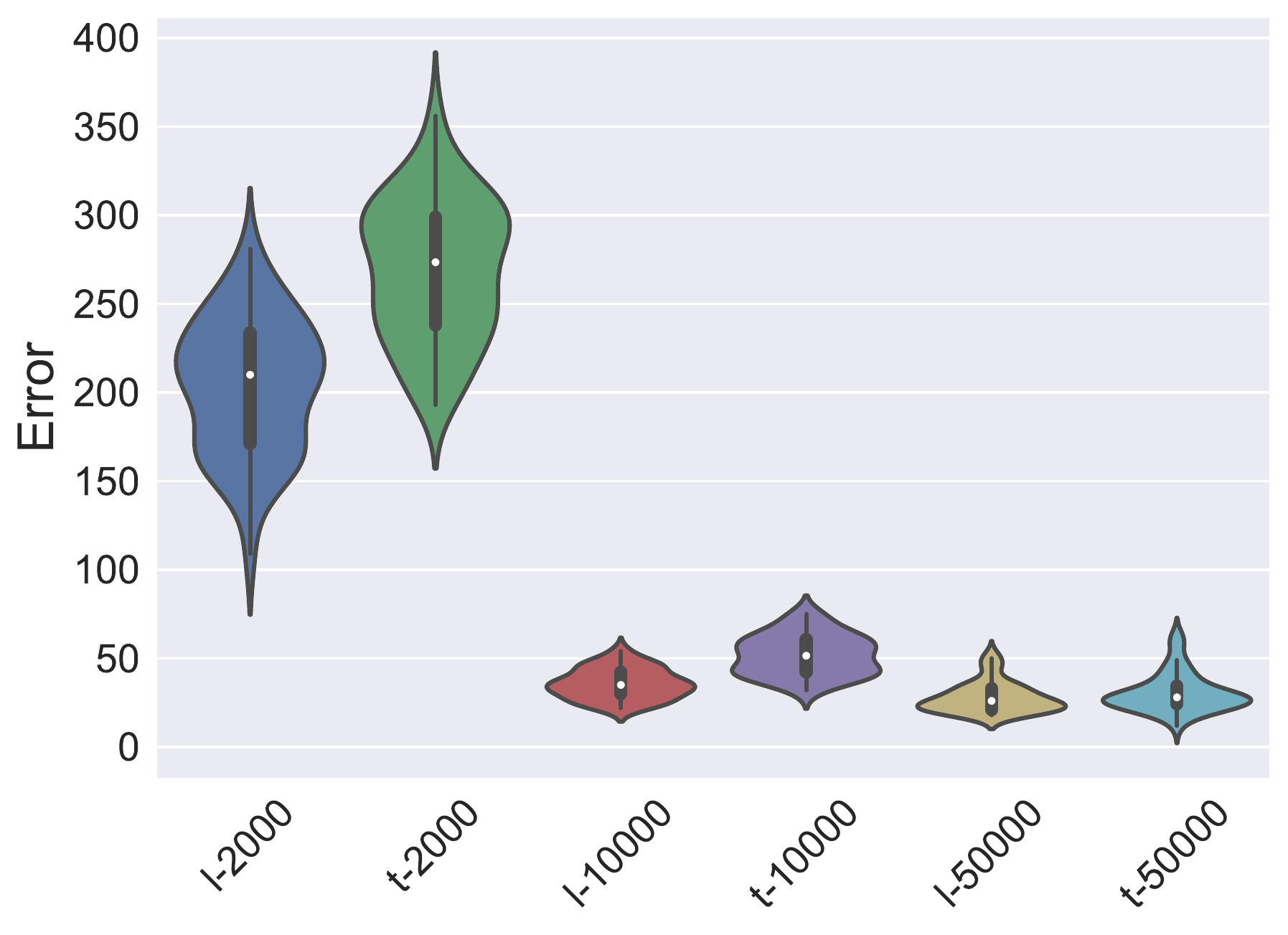}
         \caption{CLONALG.}
         \label{fig:y equals x2}
     \end{subfigure}
     
     \begin{subfigure}[t]{0.5\textwidth}
         \centering
         \includegraphics[width=\columnwidth]{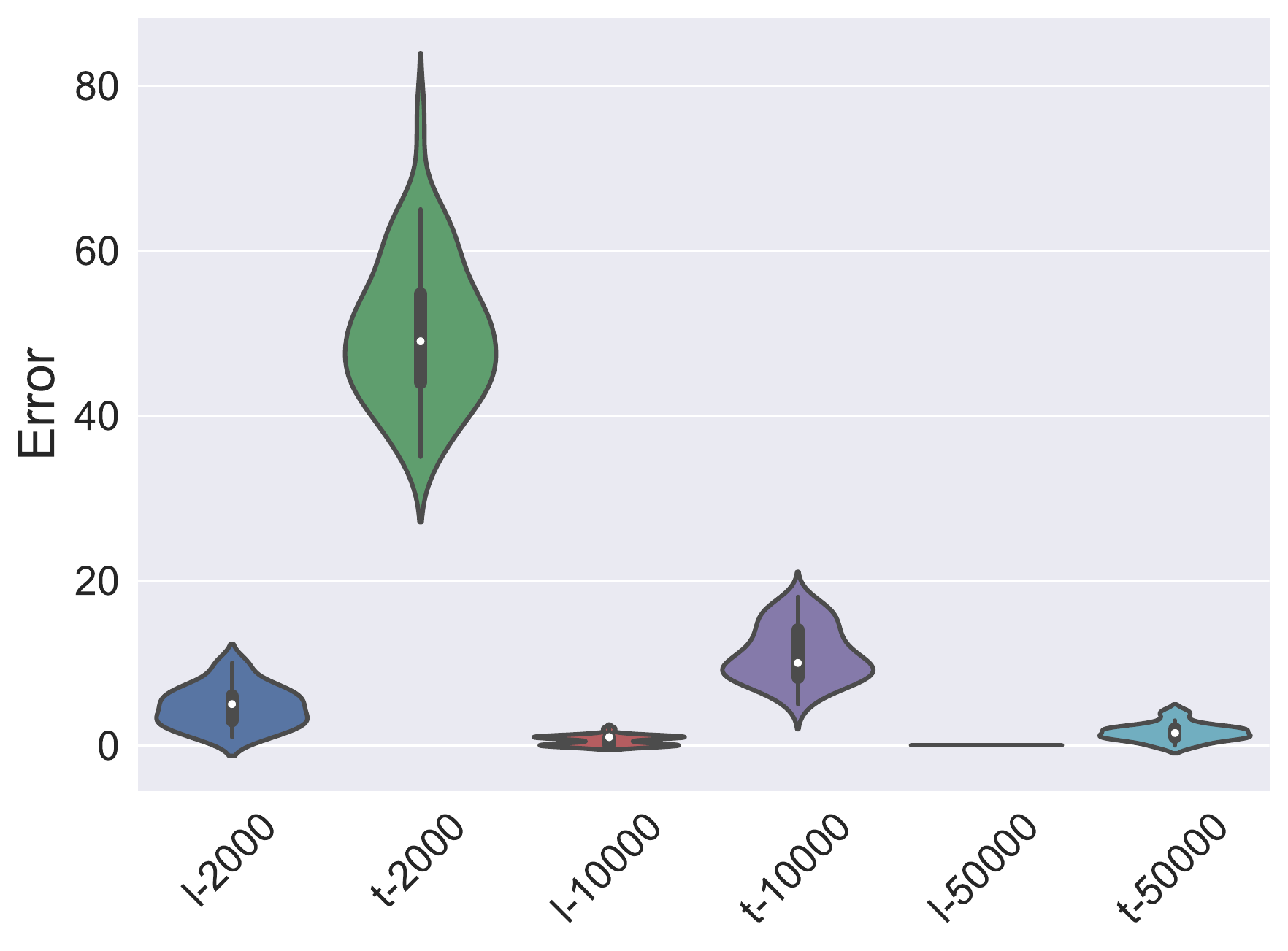}
         \caption{CMA-ES.}
         \label{fig:y equals x3}
     \end{subfigure}%
     \begin{subfigure}[t]{0.5\textwidth}
         \centering
         \includegraphics[width=\columnwidth]{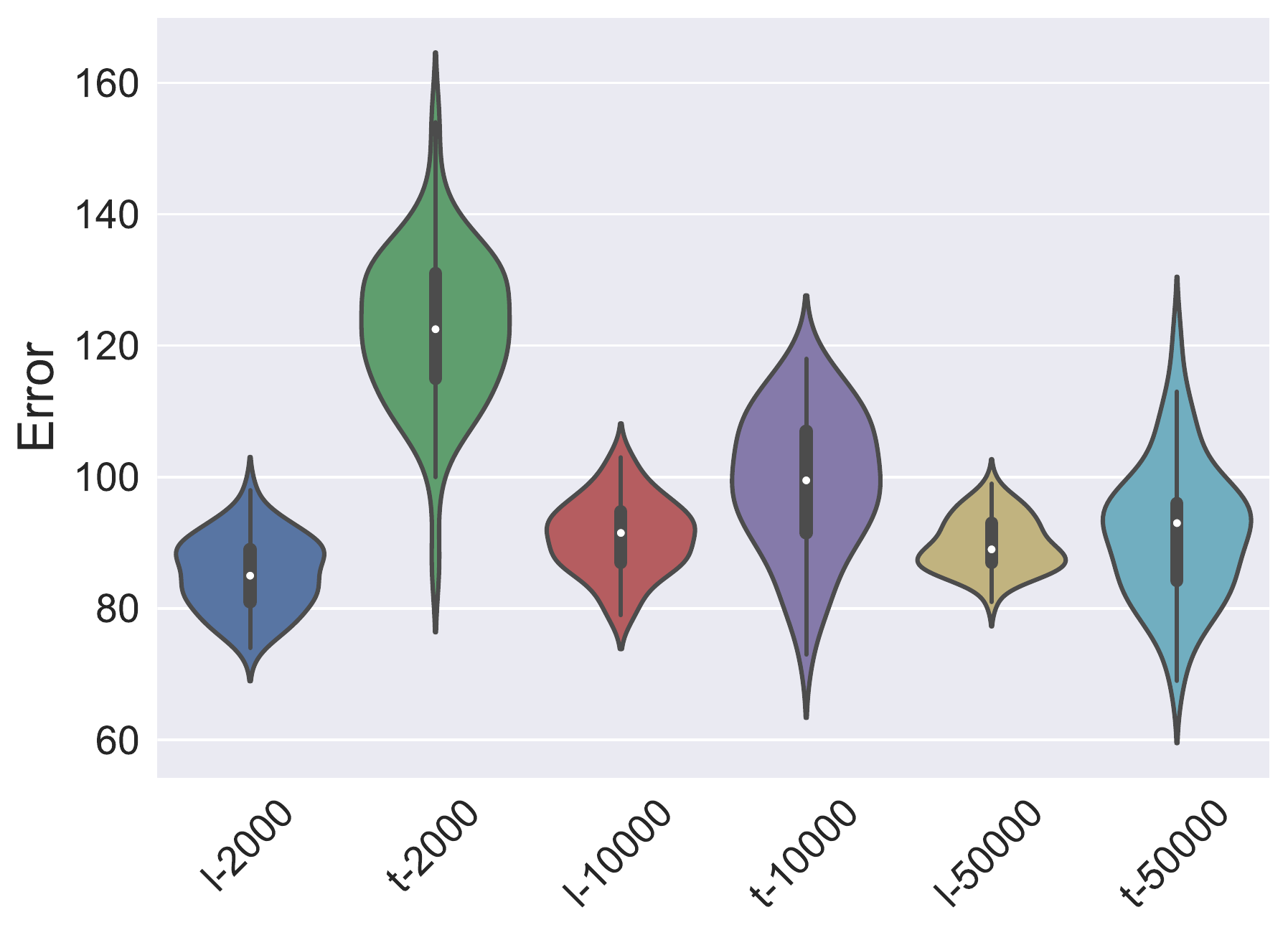}
         \caption{DE.}
         \label{fig:y equals x4}
     \end{subfigure}

     \begin{subfigure}[t]{0.5\textwidth}
         \centering
         \includegraphics[width=\columnwidth]{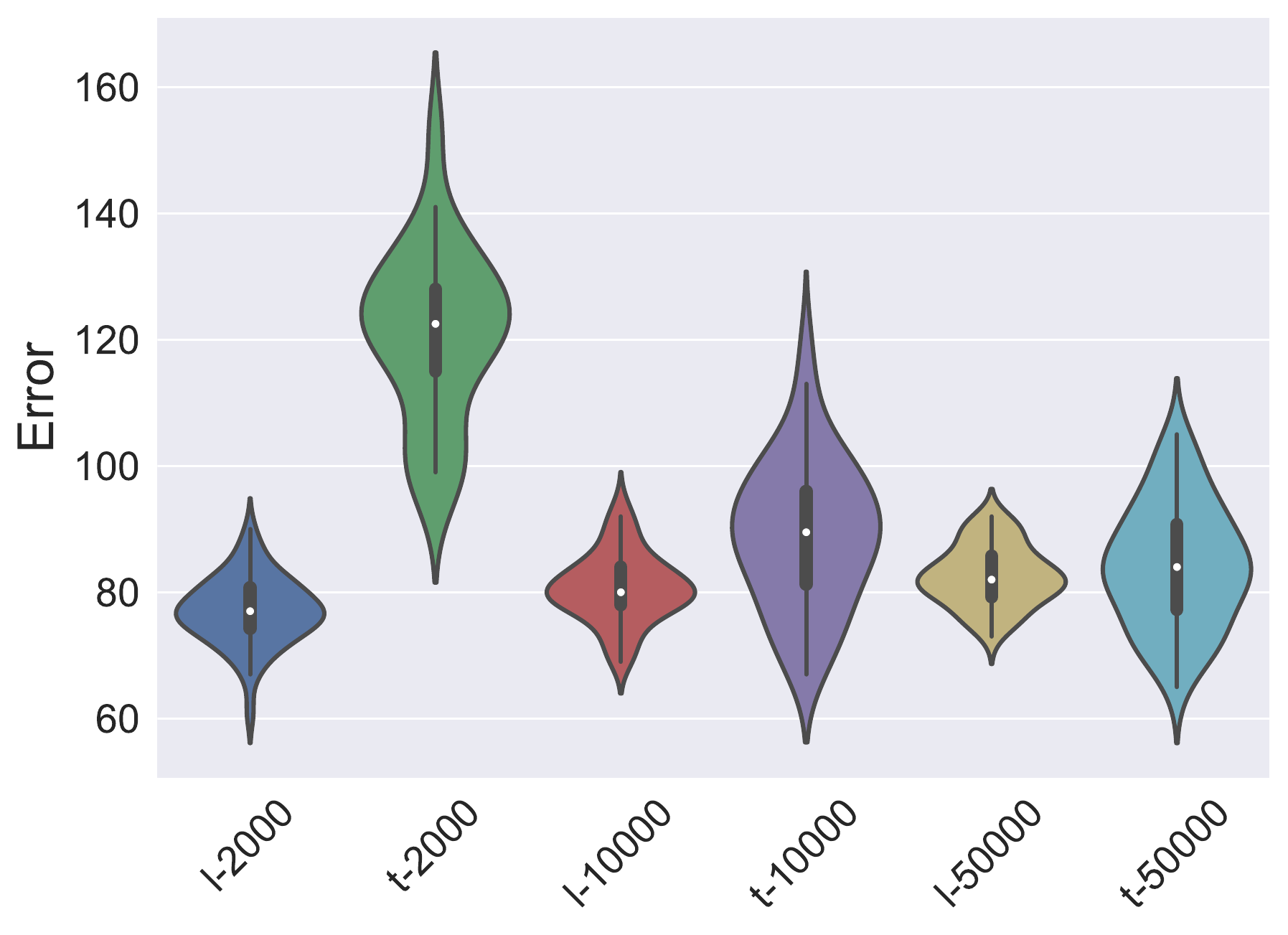}
         \caption{RW.}
         \label{fig:y equals x5}
     \end{subfigure}%
     \begin{subfigure}[t]{0.5\textwidth}
         \centering
         \includegraphics[width=\columnwidth]{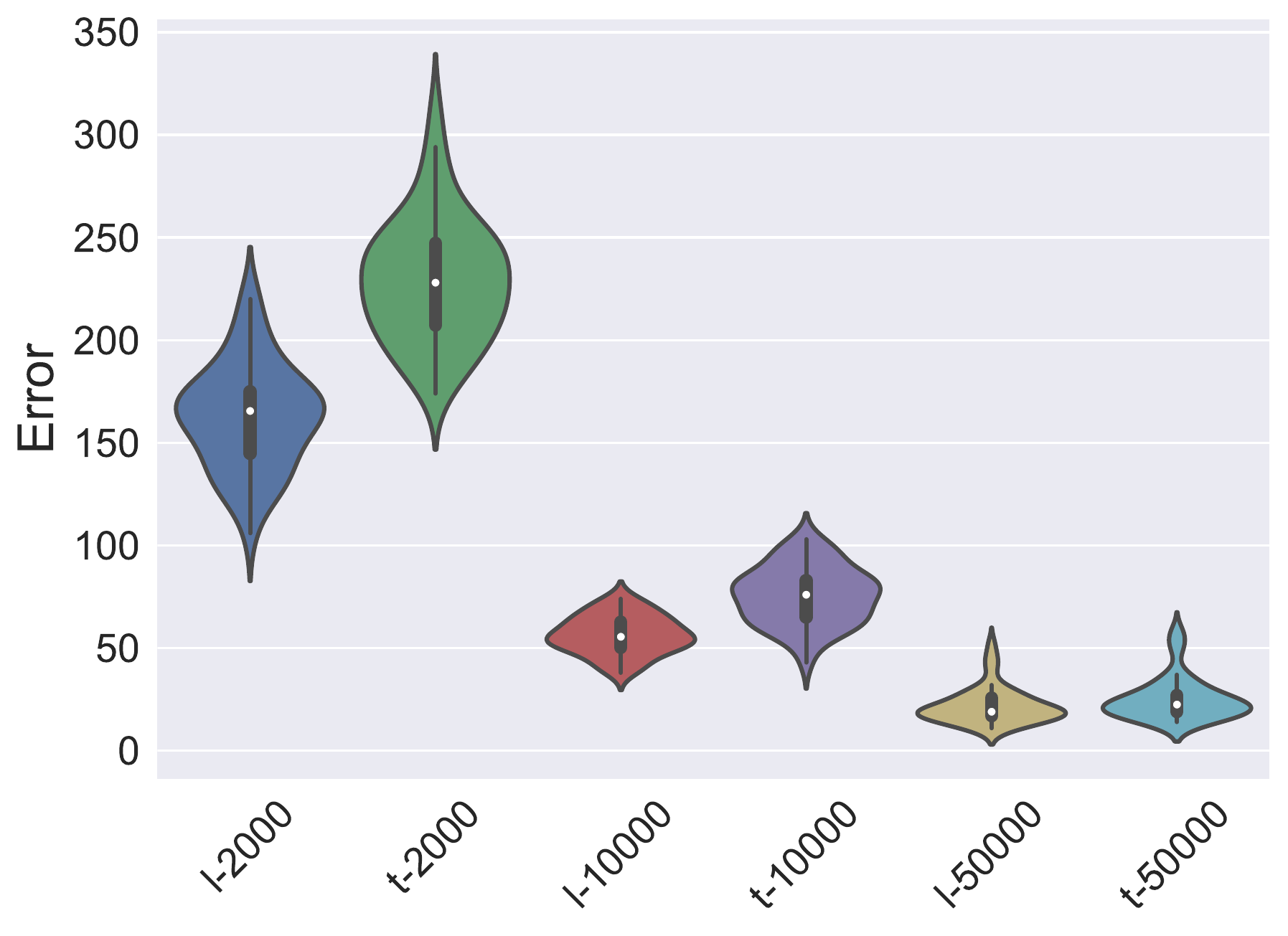}
         \caption{SST.}
         \label{fig:y equals x6}
     \end{subfigure}
        \caption{Violin plots of the results obtained for each algorithm across all executions for 1$\times$128 APUFs.}
        \label{fig:violinapuf}
\end{figure}

Figure~\ref{fig:violinapuf} shows the violin plots for all algorithms based on all the executions performed (we consider the largest PUF instance: 1$\times$128). The learning error is normalized on the scale of $[0, \ldots, 1\,000]$ to be commensurate with the test set size. The results on the training set are denoted with the prefix ``l-'' before the number of CRPs, whereas the results on the test set are denoted with the prefix ``t-''.
From the figures, we can see some degree of correlation between the training results and the test runs for some methods. This means that an algorithm performing well on the training set will find solutions that also score well on the test set. However, this is not always the case. For the smallest number of CRPs, such a conclusion cannot be drawn, as it seems that algorithms on this set easily overfit and do not generalize well. This is especially clear for CMA-ES and RW. However, for large CRPs, such correlation exists for all algorithms, and if an algorithm performs well on the learning set, it also performs well on the test set. It is also interesting to observe that for some algorithms, such as DE and RW, the errors on the training set remain similar regardless of the number of CRPs used. However, the performance on the test set improves with an increasing number of CRPs, avoiding overfitting on the training set when it is too small.

\subsection{Results for XOR APUFs}

Table~\ref{tbl:xorapuf} shows the results obtained for different XOR APUF instances. It is clear that even for the simplest instance, many algorithms have trouble achieving a low error rate. In many cases, the algorithms perform only slightly better than random estimation. Using a larger number of CRPs with these problem instances, it is necessary to achieve better results. But even this is not enough to guarantee good performance; this is best seen with the 4$\times$64 instance, where only CMA-ES obtained acceptable accuracy. The other algorithms showed no improvement even when increasing the number of CRPs. However, CMA-ES also requires a significant number of CRPs to achieve a satisfactory error score.

\begin{table}[!h]
\caption{Errors obtained by different XOR APUFs instances on the test set.}
\label{tbl:xorapuf}
\centering
\begin{tabular}{@{}llcccccc@{}}
\toprule
\multirow{2}{*}{PUF} & \multirow{2}{*}{CRP} & \multicolumn{6}{c}{Method}                                                                                                                                    \\ \cmidrule(l){3-8} 
                      &                     & \multicolumn{1}{l}{AIS} & \multicolumn{1}{l}{CLONALG} & \multicolumn{1}{l}{CMA-ES} & \multicolumn{1}{l}{DE} & \multicolumn{1}{l}{RW} & \multicolumn{1}{l}{SST} \\ \midrule
\multirow{4}{*}{4$\times$16} & 2\,000                & 468                     & 460                         & 27                        & 222                    & 409                    & 463                     \\
                      & 10\,000               & 451                     & 456                         & 0                         & 162                    & 213                    & 28                      \\
                      & 50\,000               & 166                     & 14                          & 0                         & 131                    & 160                    & 9                       \\
                      & 250\,000              & 32                      & 13                          & 0                         & 139                    & 165                    & 4                       \\ \midrule
\multirow{4}{*}{4$\times$32} & 2\,000                & 470                     & 466                         & 472                       & 468                    & 474                    & 469                     \\
                      & 10\,000               & 457                     & 474                         & 455                       & 312                    & 435                    & 464                     \\
                      & 50\,000               & 463                     & 478                         & 0                         & 292                    & 345                    & 471                     \\
                      & 250\,000              & 200                     & 52                          & 0                         & 266                    & 341                    & 11                      \\ \midrule
\multirow{4}{*}{4$\times$64} & 2\,000                & 471                     & 460                         & 462                       & 456                    & 472                    & 459                     \\
                      & 10\,000               & 429                     & 422                         & 362                       & 451                    & 431                    & 427                     \\
                      & 50\,000               & 451                     & 467                         & 471                       & 461                    & 475                    & 470                     \\
                      & 250\,000              & 454                     & 472                         & 0                         & 423                    & 469                    & 468                     \\ \bottomrule
\end{tabular}
\end{table}

\begin{figure*}[!h]
     \centering
     \begin{subfigure}[t]{0.5\textwidth}
         \centering
         \includegraphics[width=\columnwidth]{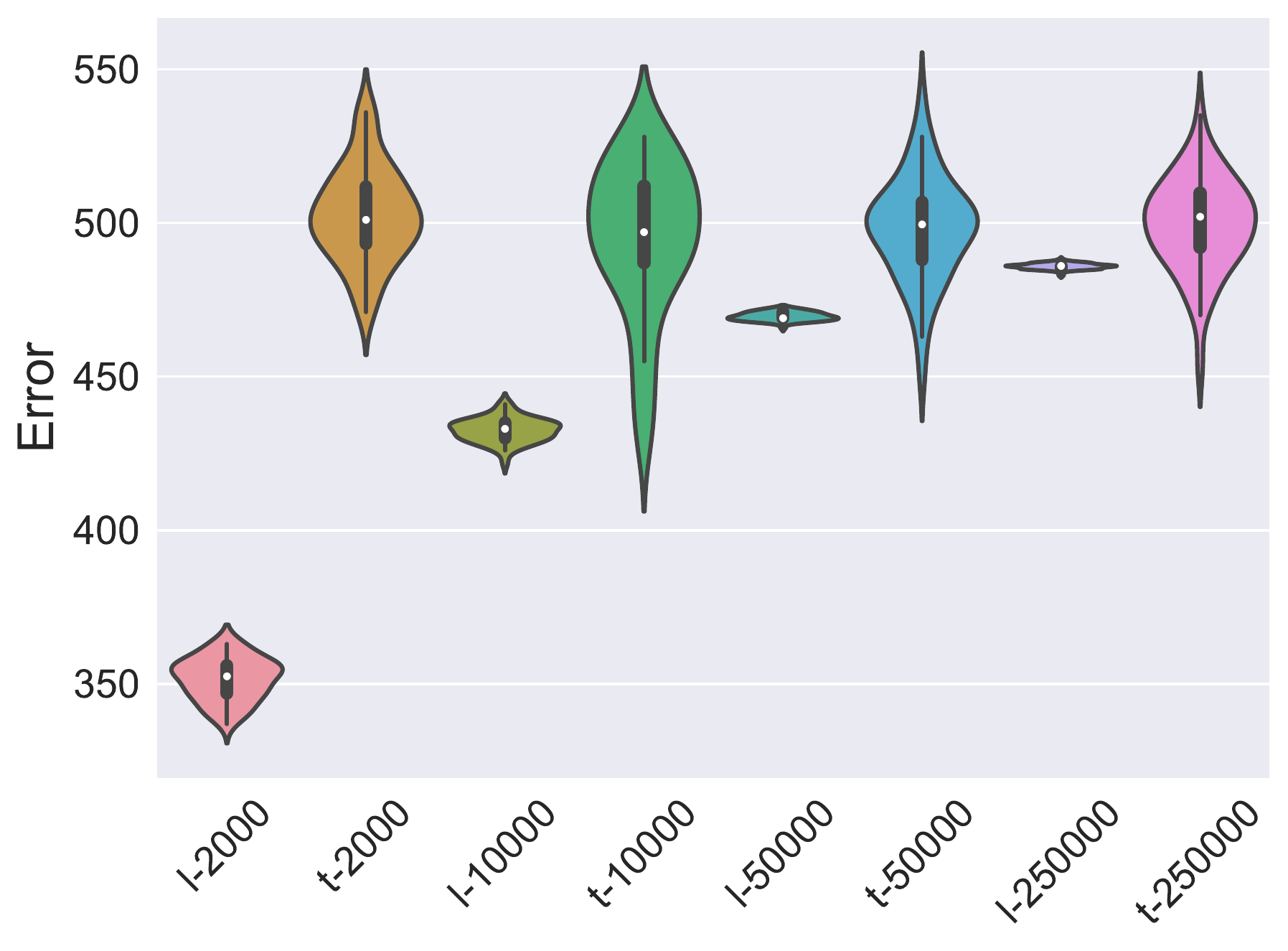}
         \caption{AIS.}
         \label{fig:y equals x7}
     \end{subfigure}%
     \begin{subfigure}[t]{0.5\textwidth}
         \centering
         \includegraphics[width=\columnwidth]{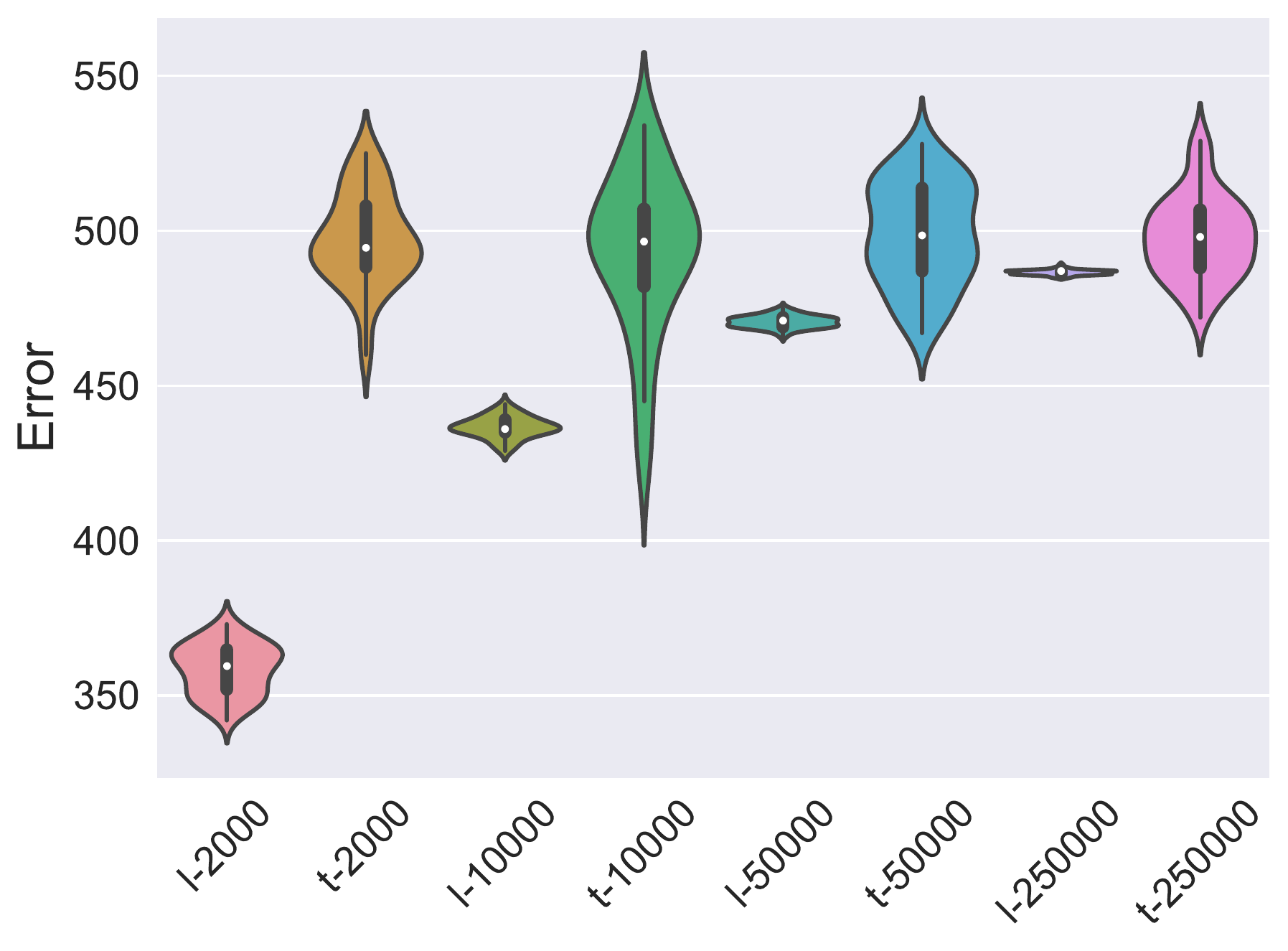}
         \caption{CLONALG.}
         \label{fig:y equals x8}
     \end{subfigure}
     
     \begin{subfigure}[t]{0.5\textwidth}
         \centering
         \includegraphics[width=\columnwidth]{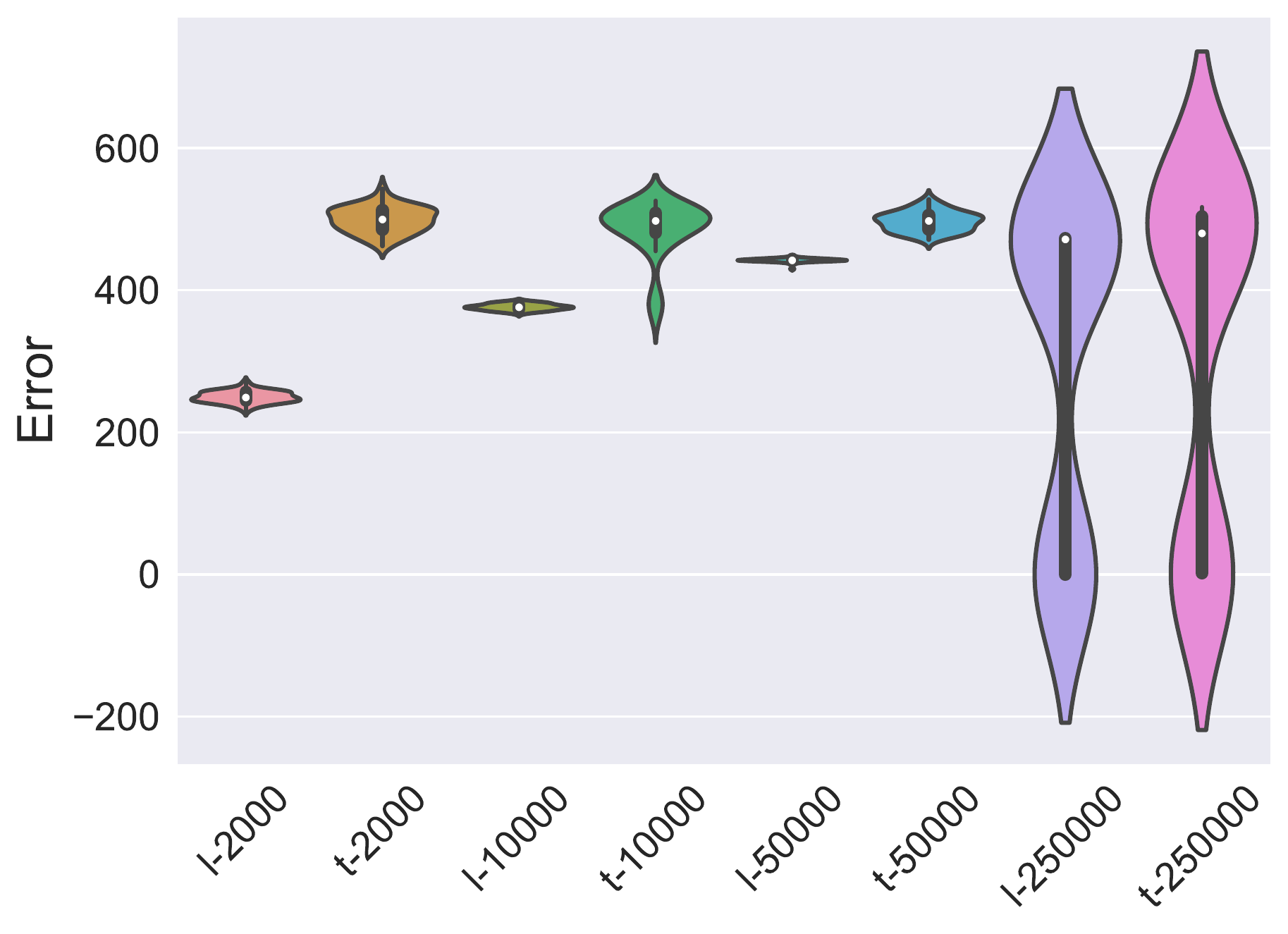}
         \caption{CMA-ES.}
         \label{fig:y equals x9}
     \end{subfigure}%
          \begin{subfigure}[t]{0.5\textwidth}
         \centering
         \includegraphics[width=\columnwidth]{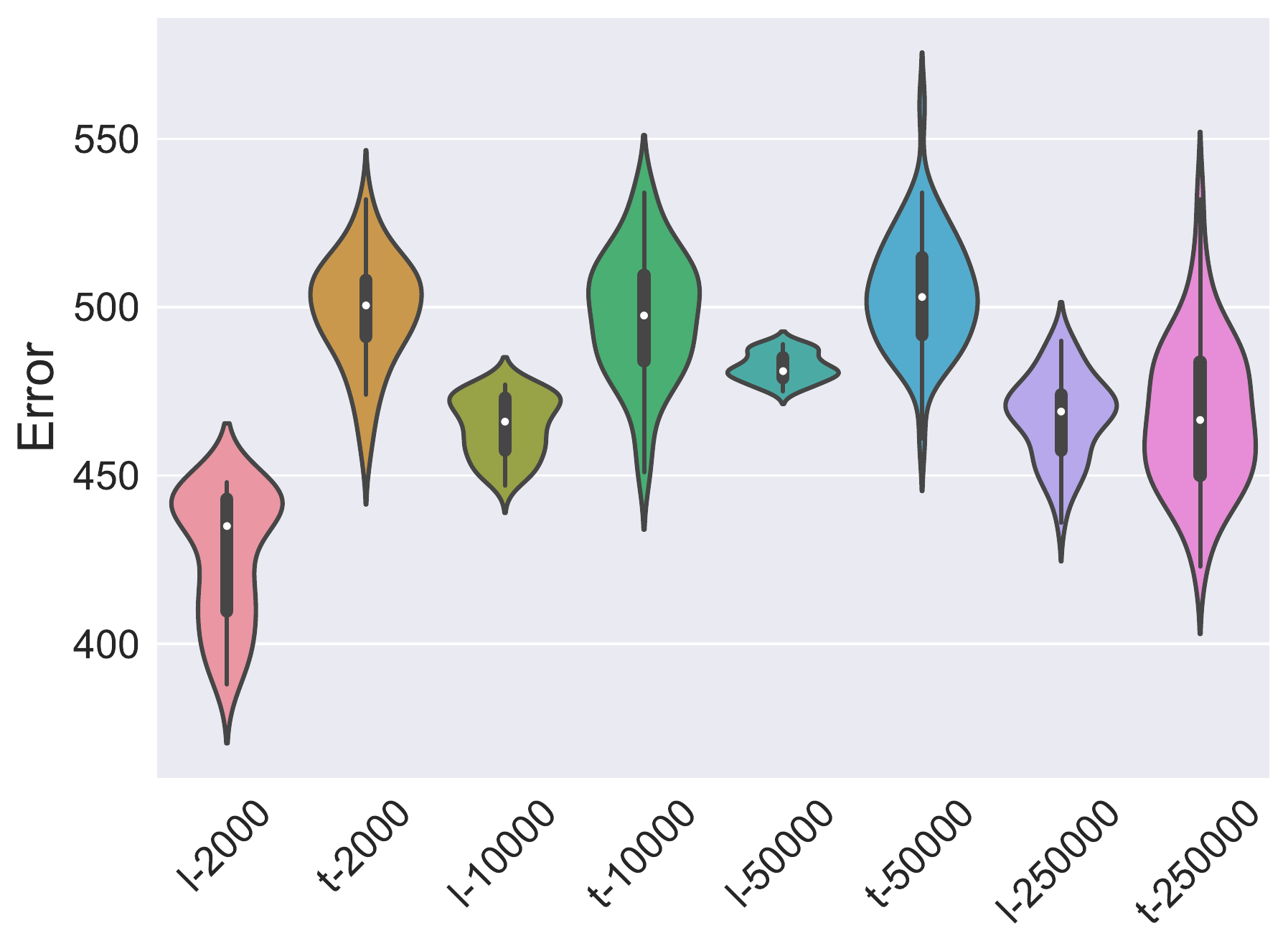}
         \caption{DE.}
         \label{fig:y equals x10}
     \end{subfigure}
     
     \begin{subfigure}[t]{0.5\textwidth}
         \centering
         \includegraphics[width=\columnwidth]{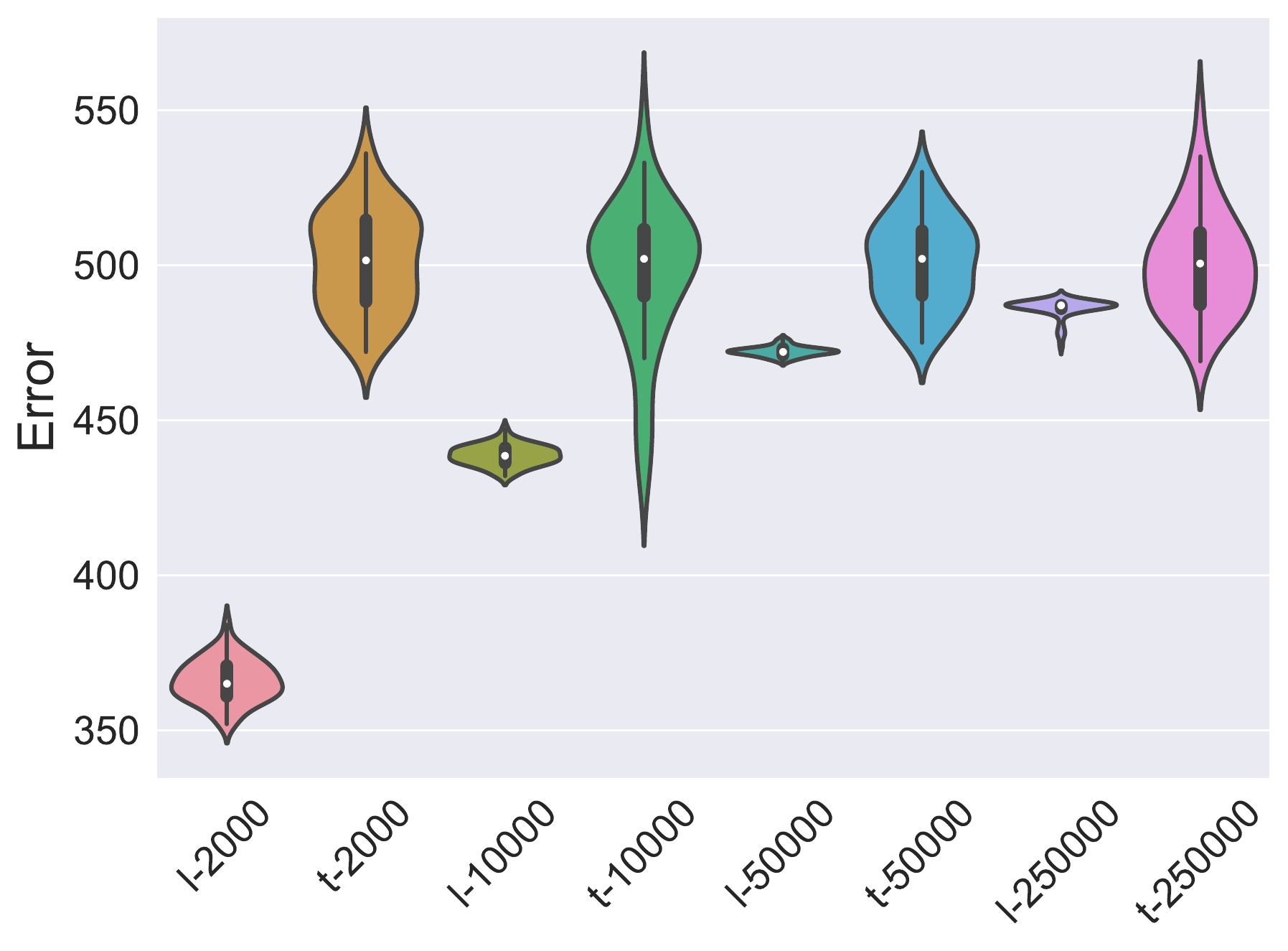}
         \caption{RW.}
         \label{fig:y equals x11}
     \end{subfigure}%
          \begin{subfigure}[t]{0.5\textwidth}
         \centering
         \includegraphics[width=\columnwidth]{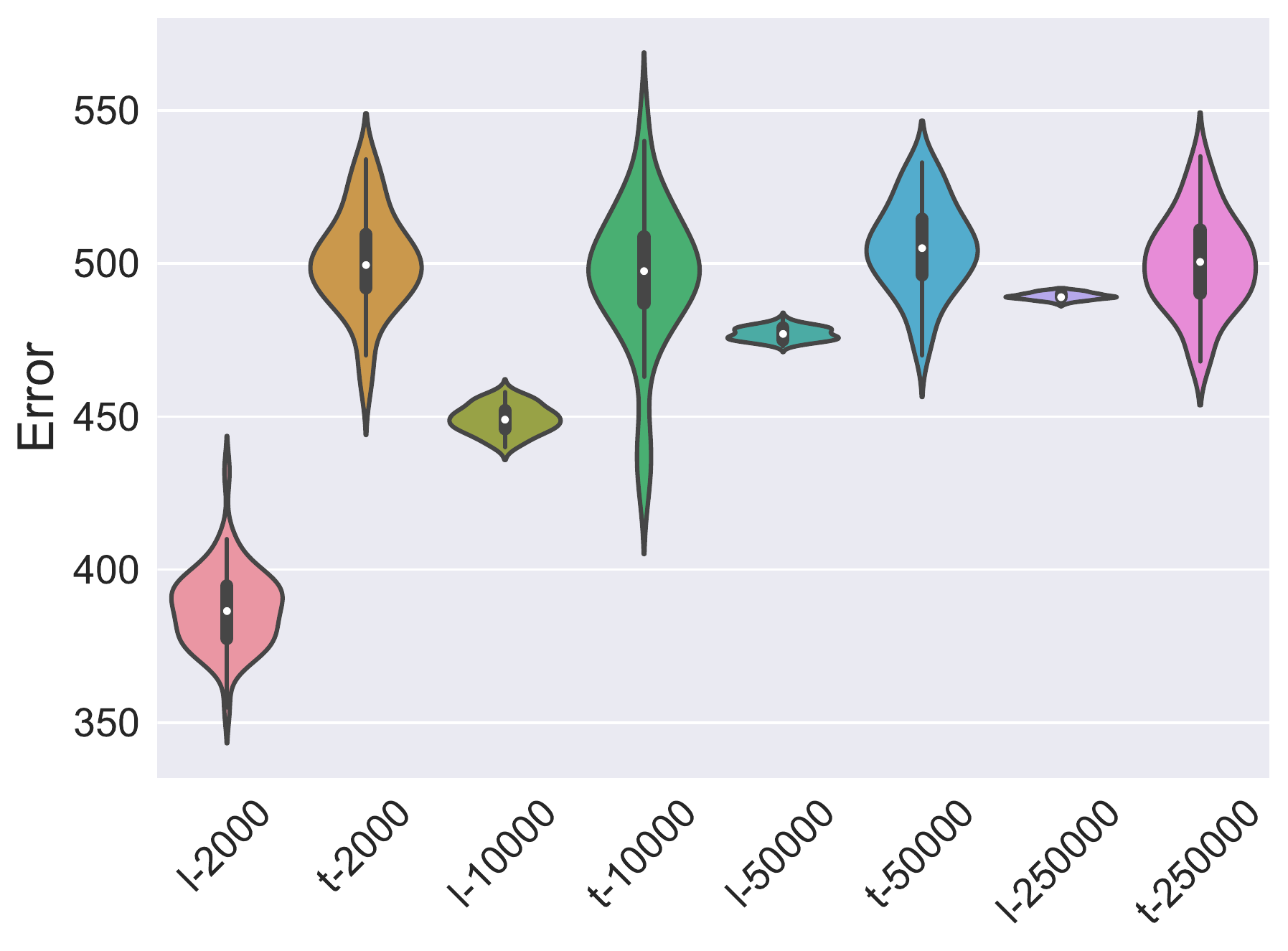}
         \caption{SST.}
         \label{fig:y equals x12}
     \end{subfigure}
        \caption{Violin plots of the results obtained for each algorithm across all executions for 4$\times$64 XOR APUFs.}
        \label{fig:violinxorapuf}
\end{figure*}

Figure~\ref{fig:violinxorapuf} shows the violin plots of the results obtained for the XOR APUF instance (4$\times$64). The figures show that all algorithms except CMA-ES achieve quite poor results. However, the situation for CMA-ES is quite interesting; for smaller CRP values, it performs similarly to the other algorithms. On the largest learning set, one can see that it may still fail to converge to a good solution, but also that some runs provide very low error on both the learning and the test set. Therefore, the violin plot is elongated and has two distinct modes. Moreover, in all cases, except for the successful CMA-ES runs, the performance on the learning set deteriorates as the number of CRPs increases, but the performance on the test set usually remains similar. This indicates that it is more difficult for the algorithms to learn the correct configuration from the given instances and that the results on the training set are misleading if not enough CRPs are used.

To further investigate the correlation between training and test sets' performance, we select CMA-ES (since it is the best performing algorithm) and calculate the correlation between its performance in the training and test sets using Spearman's rank correlation coefficient. The results of these tests are shown in Table~\ref{tbl:corr}. The table shows almost no correlation on the smallest learning sets for all PUF instances. However, as the number of CRPs increases, so does the correlation. Larger increases are observed in simple PUF instances but also in the most complex ones. For the most complex PUF instances, it can be seen that only with the largest number of CRPs there is a correlation between the performance on the test and training dataset.

\begin{table}[!h]
\caption{Results of the Spearman test for the CMA-ES algorithm}
\label{tbl:corr}
\centering
\begin{tabular}{@{}lcccc@{}}
\toprule
\multirow{2}{*}{PUF} & \multicolumn{4}{c}{CRP} \\ \cmidrule(l){2-5} 
                     & 2\,000       & 10\,000    & 50\,000    & 250\,000    \\ \midrule
4$\times$16                 & 0.157      & 0.961    & 0.962    & 0.989     \\
4$\times$32                 & -0.201     & 0.396    & 0.836    & 0.764     \\
4$\times$64                 & -0.026     & 0.224    & 0.016    & 0.803     \\ \bottomrule
\end{tabular}
\end{table}

\begin{figure}[!h]
     \centering
     \begin{subfigure}[t]{0.5\textwidth}
         \centering
         \includegraphics[width=\columnwidth]{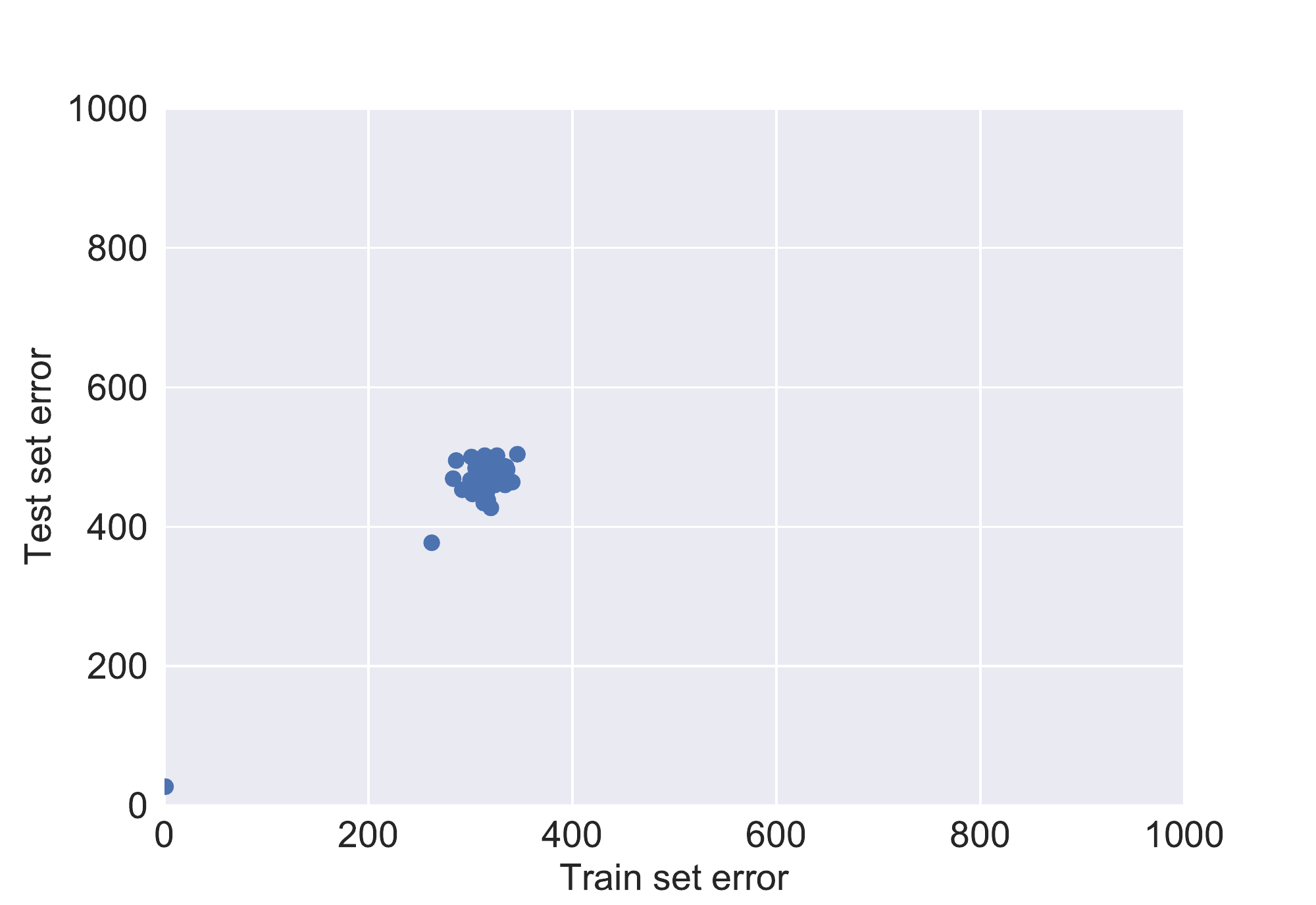}
         \caption{4$\times$16 with 2\,000 CRP.}
         \label{fig:y equals x13}
     \end{subfigure}%
     \begin{subfigure}[t]{0.5\textwidth}
         \centering
         \includegraphics[width=\columnwidth]{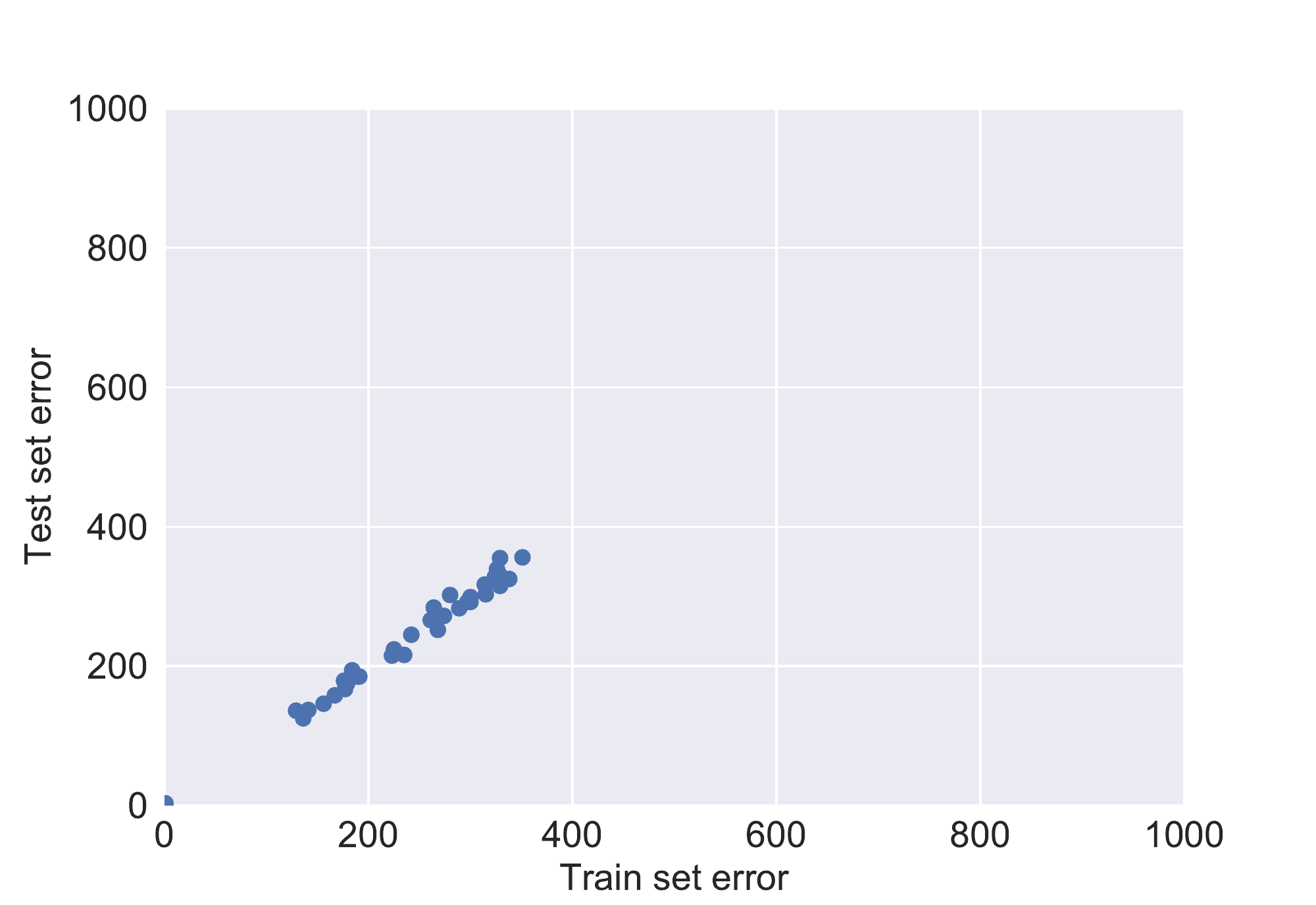}
         \caption{4$\times$16 with 250\,000 CRP.}
         \label{fig:y equals x14}
     \end{subfigure}
     
     \begin{subfigure}[t]{0.5\textwidth}
         \centering
         \includegraphics[width=\columnwidth]{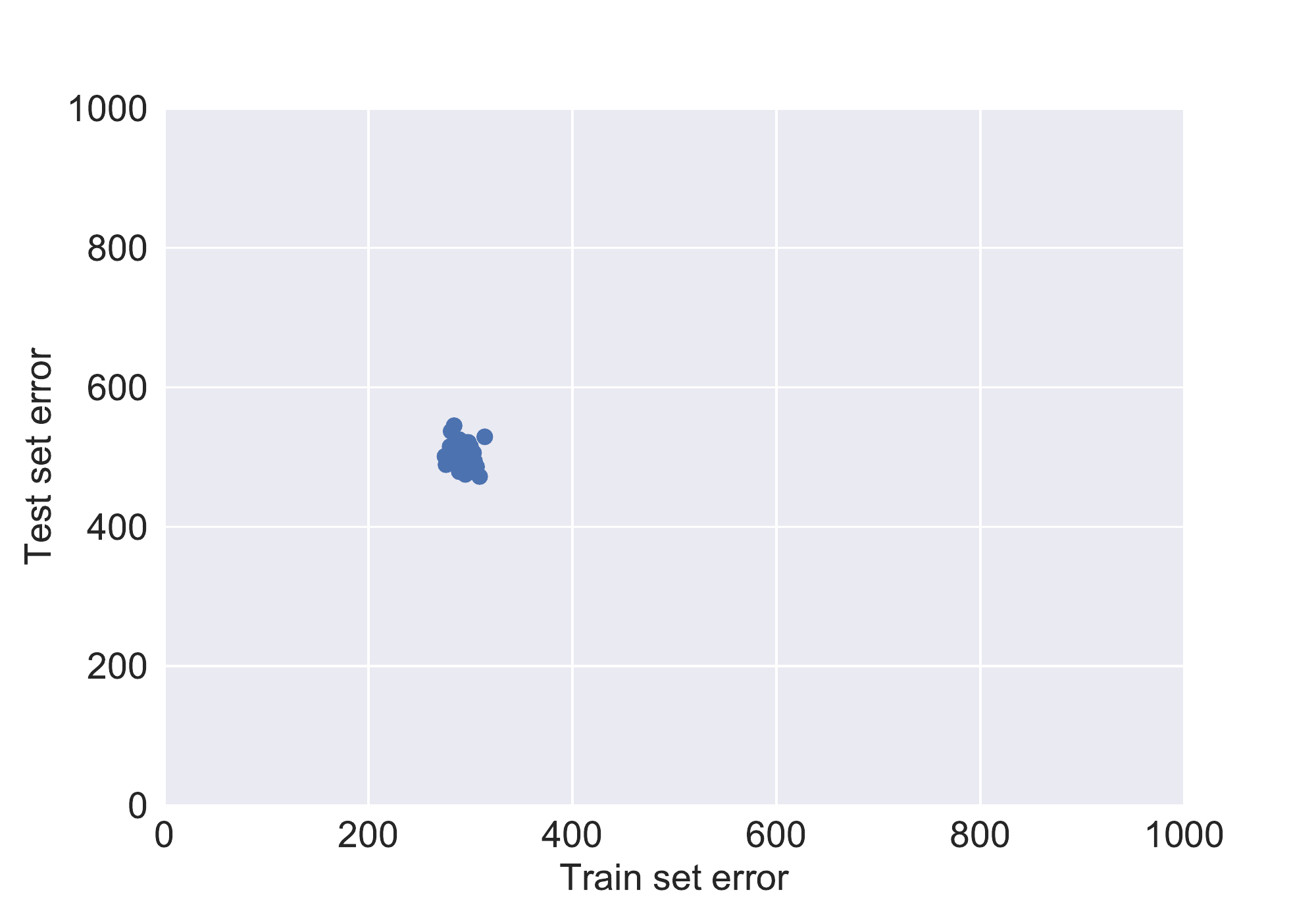}
         \caption{4$\times$32 with 2\,000 CRP.}
         \label{fig:y equals x15}
     \end{subfigure}%
     \begin{subfigure}[t]{0.5\textwidth}
         \centering
         \includegraphics[width=\columnwidth]{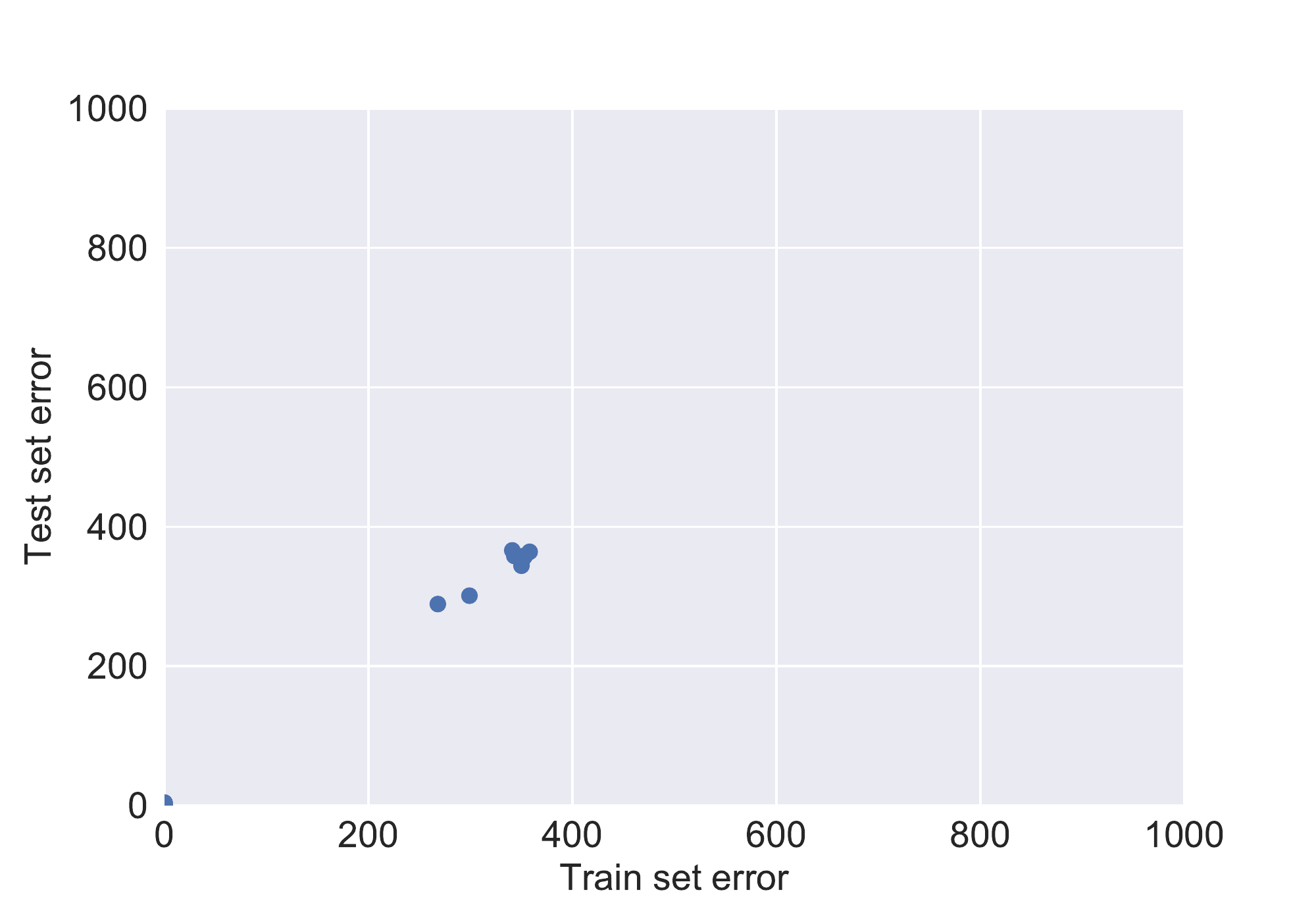}
         \caption{4$\times$32 with 250\,000 CRP.}
         \label{fig:y equals x16}
     \end{subfigure}
     
     \begin{subfigure}[t]{0.5\textwidth}
         \centering
         \includegraphics[width=\columnwidth]{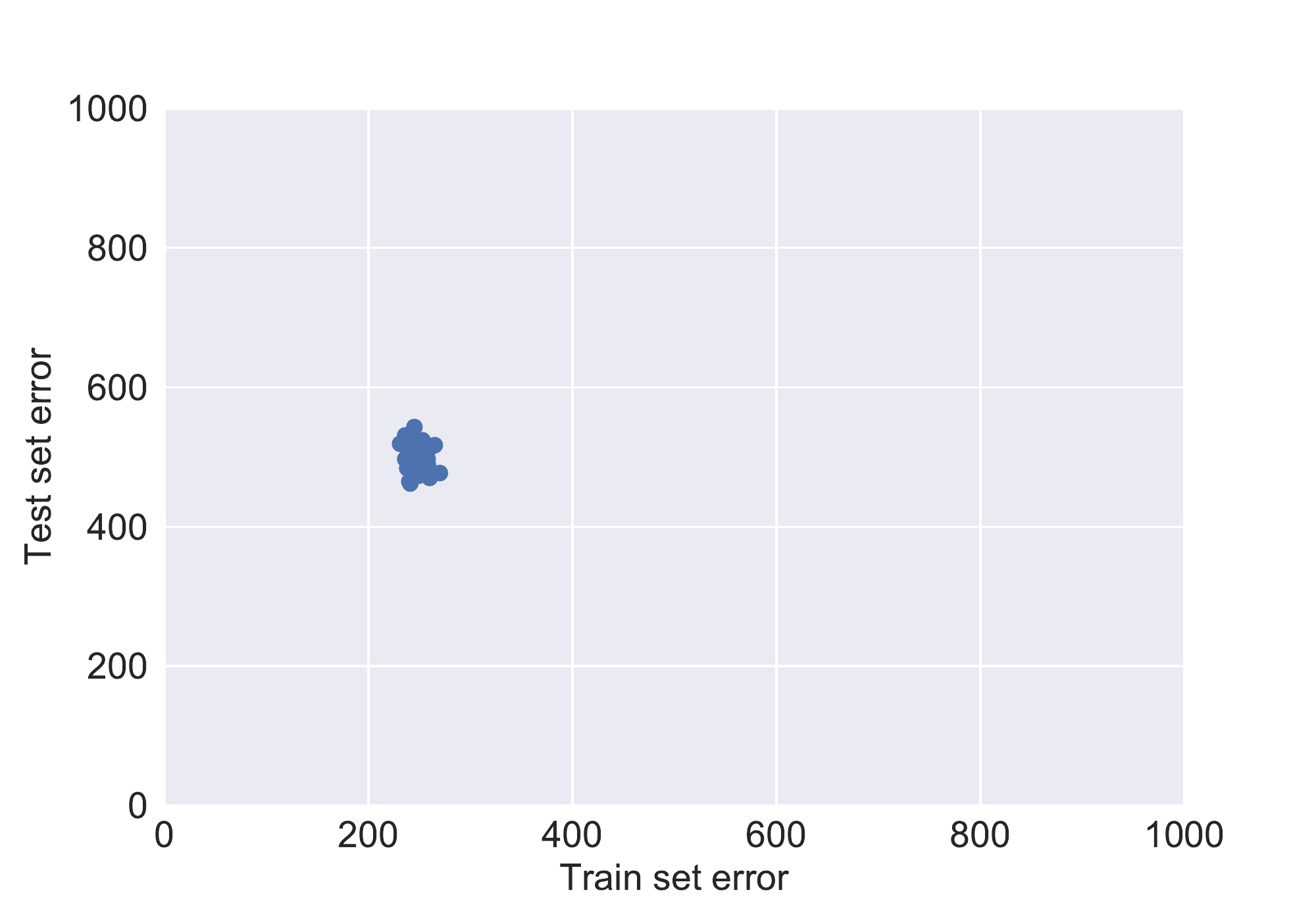}
         \caption{4$\times$64 with 2\,000 CRP.}
         \label{fig:y equals x17}
     \end{subfigure}%
     \begin{subfigure}[t]{0.5\textwidth}
         \centering
         \includegraphics[width=\columnwidth]{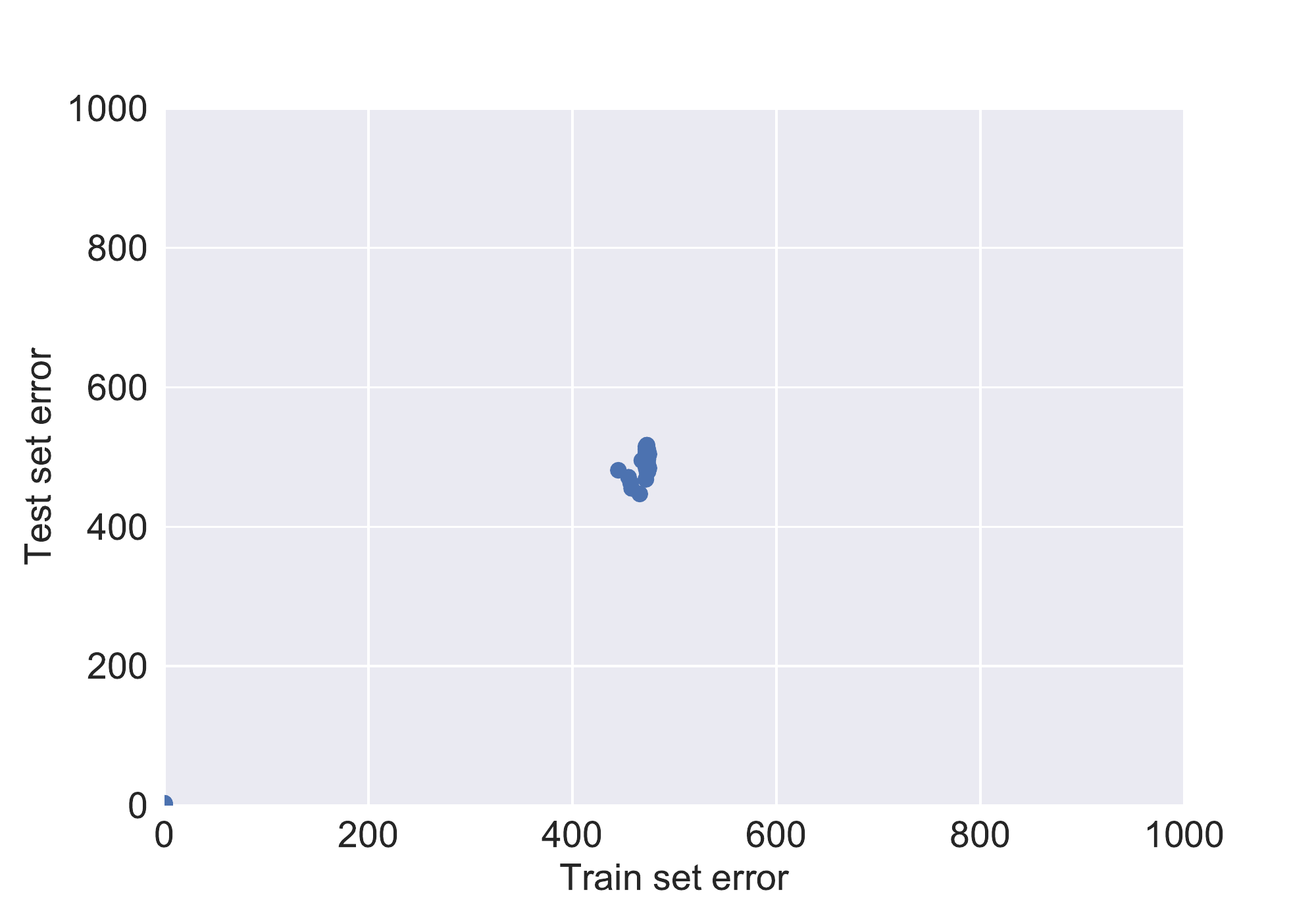}
         \caption{4$\times$64 with 250\,000 CRP.}
         \label{fig:y equals x18}
     \end{subfigure}
        \caption{Scatter plots outlining different correlation levels for CMA-ES and various XOR APUF instances.}
        \label{fig:cmaescorr}
\end{figure} 

Figure~\ref{fig:cmaescorr} shows scatter plots for CMA-ES when optimizing different PUF instances and using the least and greatest number of CRPs. The x-axis represents the error obtained on the training set, while the y-axis represents the error obtained on the test set.
In the case of the 4$\times$16 instance, it can be observed that for 2\,000 CRP, there is little correlation between the results obtained on both sets. Most of the time, an error between 200 and 400 is obtained on the learning set, but the error obtained on the test set is usually higher. In rare cases, the algorithm obtains a good configuration, which then performs well on both sets. When the number of CRPs increases, a much better correlation can be achieved. In this case, the evolved model performs similarly well in both sets. Such behavior is also observed in the two more complicated PUF instances, although the algorithm converges to good solutions less frequently. For the 4$\times$64 instance, it should be noted that, although the images look similar, the results are much better correlated for the larger number of CRPs. This is due to a number of runs obtaining a very low error, all concentrated in the lower-left corner of the plot.

\subsection{Discussion}

Based on the results presented earlier, several observations can be made. First, CMA-ES achieved the best results in all tested scenarios. For APUFs, the differences between CMA-ES and the other algorithms were not so pronounced, but for XOR APUFs, the other algorithms were inferior. Among the other considered algorithms, SST and CLONALG generally obtained the second or third best results, followed by AIS, which performed well on APUFs but not as well on XOR APUFs. DE and RW performed the worst and could not reduce the error to 0 even on the simplest instances. 

The performance of the algorithms is closely related to the number of CRPs. For APUFs, the algorithms can perform quite well even with a relatively small number of 2\,000 CRPs. However, as soon as XOR APUFs are considered, such a number of CRPs - except for the simplest problem instance - is no longer sufficient to achieve satisfactory results. Even more, if too small a number of CRPs is used, the algorithms achieve performance that is slightly better than random estimation. Therefore, it is of utmost importance to use a sufficiently large number of CRPs in the training set to obtain configurations that generalize well. 

An interesting finding from the experiments is that the results of the algorithms are trustworthy if they are equipped with a sufficient number of CRPs. This means that an algorithm that performs well in the training set will also perform similarly in the test set and vice versa. This observation makes it possible to rely on the results obtained in the training set to select the best-developed model from several runs.

\section{Conclusions and Future Work}
\label{sec:conclusions}

This work provides a systematic evaluation of six metaheuristics for the problem of modeling strong PUFs, and more precisely, Arbiter PUFs and XOR Arbiter PUFs.
The results indicate that metaheuristics can model various PUFs for the CRP-based attack. Additionally, we observe that CMA-ES is the best performing algorithm, but other metaheuristics can also successfully break the target (especially if we consider APUFs and provide enough CRPs).
We observe that the main condition for a successful attack is the size of the training set. What is more, there is a correlation between the performance on the training and test sets: if the performance on the training set is good, it will also be good for the test set.

For future work, we plan to investigate other than floating-point encodings and systematically evaluate the performance of various metaheuristics for the reliability attack. Finally, we considered here two well-known examples of strong PUFs, but the related works have many more, e.g., interpose PUFs~\cite{DBLP:journals/tches/NguyenSJMRD19}. It would be interesting to investigate the performance of metaheuristics on such, even more difficult PUFs.

\bibliographystyle{abbrv}
\bibliography{references}

\end{document}